\documentclass[conference]{IEEEtran}
\IEEEoverridecommandlockouts

\usepackage{cite}
\usepackage{amsmath}
\usepackage{amssymb}
\usepackage{url}

\usepackage{graphicx}
\DeclareGraphicsExtensions{.pdf,.jpeg,.png}

\begin{document}
%
\title{A Multi-Task Deep Learning Framework for Pose-Invariant Face Recognition and Facial Landmark Localization}
\author{\IEEEauthorblockN{Ali Ashkeyev}
\IEEEauthorblockA{Faculty of Information Technology\\
Kazakh-British Technical University\\
Almaty, Kazakhstan\\
Email: ashkeyevli@gmail.com}
\and
\IEEEauthorblockN{Zukhra Abdiakhmetova, PhD, Assoc. Prof.}
\IEEEauthorblockA{Faculty of Information Technology and Artificial Intelligence\\
Abai Kazakh National Pedagogical University\\
Almaty, Kazakhstan\\
Email: z.abdiakhmetova@abaiuniversity.edu.kz}%
}

\maketitle

\begin{abstract}

Face recognition systems often experience performance degradation under large pose variations due to changes in facial appearance and visibility of discriminative features. This study presents a multi-task deep learning framework that jointly addresses pose-invariant face recognition and facial landmark localization within a unified architecture.
  
To enable simultaneous optimization of identity recognition and facial structure learning, the original Cross-Pose Labeled Faces in the Wild (CPLFW) dataset was adapted into a multi-task learning dataset by associating facial images with identity labels and five-point landmark annotations. The proposed framework consists of two task-specific branches operating on a common facial input: a ResNet50-based branch for identity recognition and an EfficientNetV2B0-based branch with spatial attention for facial landmark localization. The two objectives are jointly optimized using a weighted multi-task loss, allowing the model to learn complementary semantic and spatial facial representations.
  
The proposed framework was evaluated on the prepared CPLFW-based multi-task dataset. Unlike the original CPLFW benchmark, which focuses primarily on identity recognition, the proposed dataset adaptation incorporates five-point facial landmark annotations to enable joint recognition and localization learning. Experimental results demonstrate that the model achieves a validation recognition accuracy of 81.58\% and a facial landmark localization mean squared error of 96.38.
The results indicate that combining identity classification and landmark regression within a single optimization framework provides an effective approach for learning robust facial representations under cross-pose conditions.
  
The proposed method demonstrates the potential of multi-task learning and attention-based feature refinement for unconstrained facial analysis.
  
  \end{abstract}

  \begin{IEEEkeywords}
  face recognition, facial landmark localization, multi-task learning,
  pose invariance, spatial attention.
  \end{IEEEkeywords}

  \IEEEpeerreviewmaketitle

  \section{Introduction} 
   Face recognition has become an important biometric technology for identity verification, access control, surveillance, and security applications. Early face recognition systems primarily relied on handcrafted representations and statistical dimensionality-reduction techniques. Principal Component Analysis (PCA), for example, formed the basis of the Eigenfaces approach, in which facial images are represented in a lower-dimensional eigenspace \cite{turk1991eigenfaces}. Other approaches, including Linear Discriminant Analysis (LDA) and Fisherface-based methods\cite{belhumeur1997fisherfaces}, Local Binary Patterns (LBP)\cite{ahonen2004lbp}, and Gabor-based representations\cite{4811625}, further improved facial feature extraction. Nevertheless, traditional methods remain sensitive to variations in illumination, expression, occlusion, and particularly head pose, which limits their reliability in unconstrained environments.

Convolutional Neural Networks (CNNs) improved face recognition technologies. This approach  allows for hierarchical and discriminative facial representations learn from image data. There are several examples of CNN such as DeepFace, FaceNet and ArcFace. Introduction of DeepFace combines deep representations with facial alignment\cite{taigman2014deepface}. The embedding-based approach where distances between facial representations correspond to identity similarity was shown in FaceNet\cite{schroff2015facenet}, while ArcFace shows how differences between facial features can be learned through an additive angular margin loss\cite{deng2019arcface}. The wider context of representation-learning methods can be found in following works. According to a review by Masi et al., the main components of deep face recognition are face preprocessing, network architectures, and training objectives \cite{masi2018survey}, while Wan et al. investigated the combination of CNN and subspace learning\cite{wan2017subspace}.

Changes in head orientation can significantly alter both the appearance of a face and the visibility of its key components. Under non-frontal poses, facial regions such as the eyes, nose, and mouth may be distorted or partially hidden, increasing intra-class variation between images of the same individual. The problem becomes more difficult when pose variations occur together with illumination changes, low image resolution, facial expressions, or partial occlusion. This challenge is particularly important in real-world applications such as surveillance and biometric authentication, where frontal facial images cannot always be guaranteed.

In previous research works multi-task learning (MTL) showed good performance on jointly addressing face detection, landmark localization, pose estimation, gender classification, and other facial analysis tasks\cite{ranjan2017hyperface,hsieh2017multitask,zhao2021fersnet,wang2019awsnet}. This is because MTL optimizes facial analysis tasks within a unified framework. For example, it was combined face detection, landmark localization, pose estimation, and gender recognition within a single deep multi-task architecture in HyperFace approach\cite{ranjan2017hyperface}. Structural information provided by facial landmark localization helps to solve face recognition problems in dynamic environment, because landmarks can describe spatial positions of key facial features and help to recognize facial geometry.

In pose variations, visibility of key facial features can be significantly changed, while attention mechanisms address these challenges by emphasizing key regions instead of treating all parts of an image equally. The research work that investigated spatial and channel-wise mechanisms to improve feature representation and robustness under changing conditions\cite{yan2023attention,liu2025attention,zhao2021occlusion} motivated to include spatial attention mechanism to proposed work.

Multi-task facial analysis frameworks usually merge face recognition branch with additional features like identification of gender, age or facial expression. This study proposes a MTL framework for pose-invariant face recognition and facial landmark localization, because optimized identity recognition and facial landmark localization for pose invariance are not explored fully. Proposed architecture uses ResNet50 for the identity recognition branch and EfficientNetV2B0 for facial landmark localization, while spatial attention mechanism was used to emphasize informative facial regions to help task-specific components learn discriminative identity representations and facial landmark positions. To show good merged performance of identity classification and landmark regression in framework it combined identity-related semantic information with structural facial information. For training and evaluating it was used augmented dataset derived from the Cross-Pose Labeled Faces in the Wild (CPLFW) dataset. The dataset contains photographs of people, mostly famous figures, from different angles and facial poses, and is labeled with their names, which well aligns dataset with the objective of this study\cite{zheng2018cplfw}.

The main contributions of this work are: (1) the development of a multi-task framework that jointly learns facial identity recognition and landmark localization; (2) the integration of spatial attention to emphasize informative facial regions under pose variations; and (3) the experimental evaluation of a dual-branch deep learning architecture designed to learn both semantic identity and structural facial representations from cross-pose facial data. The experimental results demonstrate that both tasks can be learned within the proposed framework, achieving a validation recognition accuracy of 81.58\% and a validation landmark localization mean squared error (MSE) of 96.38~$\mathrm{pixel}^2$.

The structure of this research paper is organized in following way. Proposed methodology, including related methods, architecture, dataset preparation, and training strategy was illustrated in Section \ref{sec:methods} called Methods, while Section \ref{sec:results} shows experimental results and Section \ref{sec:discussion} discusses these findings and limitations. The conclusion of this study and future research are shown in Section \ref{sec:conclusion}.

  \section{Methods}
  \label{sec:methods}

  \subsection{Related Methods and Design Motivation}
  \label{subsec:related}

  \subsubsection{Traditional Face Recognition Approaches}
  
Early face recognition systems mainly relied on handcrafted feature extraction and statistical representation techniques. The Eigenfaces method introduced by Turk and Pentland represented facial images using Principal Component Analysis (PCA), reducing image dimensionality while preserving dominant facial variations \cite{turk1991eigenfaces}. Linear Discriminant Analysis (LDA) and Fisherface-based methods further improved recognition performance by maximizing between-class separability and minimizing within-class variations \cite{belhumeur1997fisherfaces}. Local Binary Patterns (LBP) and texture-based descriptors were also widely investigated because of their computational efficiency and robustness to illumination changes\cite{ahonen2004lbp}. Before the widespread use of learned deep representations, Brunelli and Poggio illustrated the differences between explicitly measuring facial structure and comparing image patterns appearance \cite{brunelli1993face}. They used image characteristics for template matching and compared it with face recognition based on geometric features.
  
Although traditional approaches achieved promising results under controlled conditions, they generally depend on manually designed features and show limited robustness against illumination changes, facial expressions, occlusions, and large pose variations. These limitations motivated the transition toward deep-learning-based feature learning.

  \subsubsection{Deep Learning-Based Face Recognition}
  
  The introduction of Convolutional Neural Networks (CNNs) improved face recognition systems and allow automatically extracting hierarchical and discriminative representations. Combination of deep convolutional features and alignment of face in DeepFace showed the effective way for large-scale face verification \cite{taigman2014deepface}. In FaceNet  identity similarity was measured by using distances between embeddings. This is an embedding-based approach where images of faces are mapped into a compact feature space\cite{schroff2015facenet}. The provided context from research works that adapted pretrained recognition models was useful for this study, even if their experimental settings differ. Authors demonstrated a retrained version of a ResNet-50-based VGGFace recognition system \cite{hasan2021retraining}, while another study shows a comparison of ArcFace, FaceNet, and FaceNet512 within the DeepFace framework \cite{firmansyah2023comparison}.

  Afterward, there were improvements in the discriminative capability of facial representations. VGGFace2 is one of the examples of large-scale dataset that showed why it is important to contain variations in pose and age for learning robust facial representations  \cite{cao2018vggface2}. According to the study of Horiguchi and team under good conditions, such as sufficient training examples are available per class, softmax-based features can be competitive with distance-metric-learning features \cite{horiguchi2020softmax}. Also from research study of center loss, we can observe that center loss encourages features from the same identity to remain close to their class center\cite{wen2019centerloss}. For CosFace, the additive cosine margin was used\cite{wang2018cosface}, which differs from approach of margin-mining softmax, where authors used margin-based formulation \cite{lee2022angular}. These examples show how training objective is important for representation learning and they provide alternatives to compare with the categorical cross-entropy used in this study.

  Changes in head orientation can significantly change facial appearance and reduce the visibility of key face regions; therefore, recognition under significant pose variations is one of the critical problems.

  \subsubsection{Pose Variation and Facial Landmark Localization}
  
Pose variation introduces substantial intra-class differences because important facial components may become distorted or partially invisible. This problem is particularly important in unconstrained applications such as surveillance and biometric authentication, where frontal facial images cannot always be guaranteed. Settings in different research works share common ground with cross-pose recognition, but have different challenges, so good performance under one condition cannot be taken as evidence of robustness under all others. Research of Hsu et al. compares loss functions under pose, age, occlusion, and resolution changes \cite{hsu2020crossfactor}. Another study of Deng and Zafeiriou examines the performance of ArcFace for disguised face recognition \cite{deng2019disguised}.

Facial landmark localization provides complementary structural information by estimating the spatial positions of important facial components. Instead of treating identity recognition and landmark estimation as independent tasks, multi-task learning enables related facial objectives to be optimized jointly.

Previous approaches have explored combining detection and landmark localization within unified frameworks. Multi-task cascaded convolutional networks jointly perform face detection and landmark localization through a sequence of deep networks \cite{zhang2016mtcnn}. HyperFace extended this concept by combining face detection, landmark localization, pose estimation, and gender recognition within a unified multi-task architecture \cite{ranjan2017hyperface}.

  \subsubsection{Attention-Based Facial Representation Learning}
  
  By increasing the weight of key facial regions and decreasing background noise and other regions, attention mechanisms begin to play a key role. Concept Squeeze-and-Excitation Networks which also called adaptive feature recalibration improved convolutional representations. It regulated the importance of different channels\cite{hu2018senet}. Next step was combination of channel and spatial attention which even more fine-tuned feature representation. This approach called the Convolutional Block Attention Module (CBAM)\cite{woo2018cbam}.

  Attention mechanisms were explored in recent facial analysis studies to improve its recognition accuracy especially under dynamic environment. For example Yan and colleagues explored attention-based facial expression recognition network\cite{yan2023attention}, while Liu and team looked how to improve attention mechanisms  for face recognition representations   \cite{liu2025attention}. Also, for situations when face is partially covered Zhao et al. used attention-based refinement \cite{zhao2021occlusion}. Attention is the part of different research works about facial analysis tasks. Next studies are more about detection and expression analysis rather than identity recognition, but they are helpful in study of selective facial feature processing. For example, an attention mechanism was used to guide component localization in face detection \cite{han2009attention}, while another study investigated lightweight fused attention for facial expression recognition \cite{yu2022lightweight}.
  
  Focus of models on key facial features during pose variations or partially covering effect of the face can be improved by attention mechanisms, this was demonstrated in the studies above.

  \subsubsection{Multi-Task Learning for Facial Analysis}
  
  Multi-task learning (MTL) provides a framework for jointly optimizing related
  learning objectives by exploiting complementary information between tasks. In facial
  analysis, MTL has been applied to combine identity recognition with additional tasks
  such as attribute estimation, expression recognition, and structural localization.
  
  Previous studies investigated multi-task learning for face identification and
  attribute estimation
  \cite{hsieh2017multitask}. Facial expression recognition and synthesis were also
  formulated as multi-task problems using selective feature sharing mechanisms
  \cite{zhao2021fersnet}. Wang et al. proposed an adaptively weighted sharing network
  for jointly learning emotion recognition and facial action-unit detection
  \cite{wang2019awsnet}. The study of Zhang et al. used joint learning and feature integration, they investigated face detection using multi-task learning and multi-layer feature fusion \cite{zhang2017multilayer}, while this study uses a different task combination: identity classification and five-point facial landmark regression.
  
  These studies motivate the proposed framework. However, many existing approaches
  combine identity recognition with demographic attributes or expression-related tasks.
  In contrast, this work focuses on the complementary relationship between identity
  classification and facial landmark localization under cross-pose conditions.

  \subsubsection{Design Motivation}
  
  Face recognition needs identity-related representations to separate different people, while landmark localization records facial key features to describe facial geometry. Combination of these features gives valuable guidance for learning.

  The proposed multi-task model merges two tasks into single framework and optimizes identity classification,  landmark regression. One branch used for recognition by learning identity-related representations, while second one for landmarks by learning facial geometry improved by spatial attention.

  \subsection{Proposed Multi-Task Architecture}
  \label{subsec:architecture}

The proposed framework performs identity recognition and facial landmark localization simultaneously from a common red-green-blue (RGB) facial image. Each input image is represented as a $250 \times 250 \times 3$ tensor and is processed by two task-specific branches.
  
The identity recognition branch employs ResNet50, while the landmark localization branch uses EfficientNetV2B0 followed by a spatial-attention module. Both branches are optimized jointly using a weighted multi-task objective.
  
\begin{figure*}[t]
  \centering
  \includegraphics[width=\textwidth]{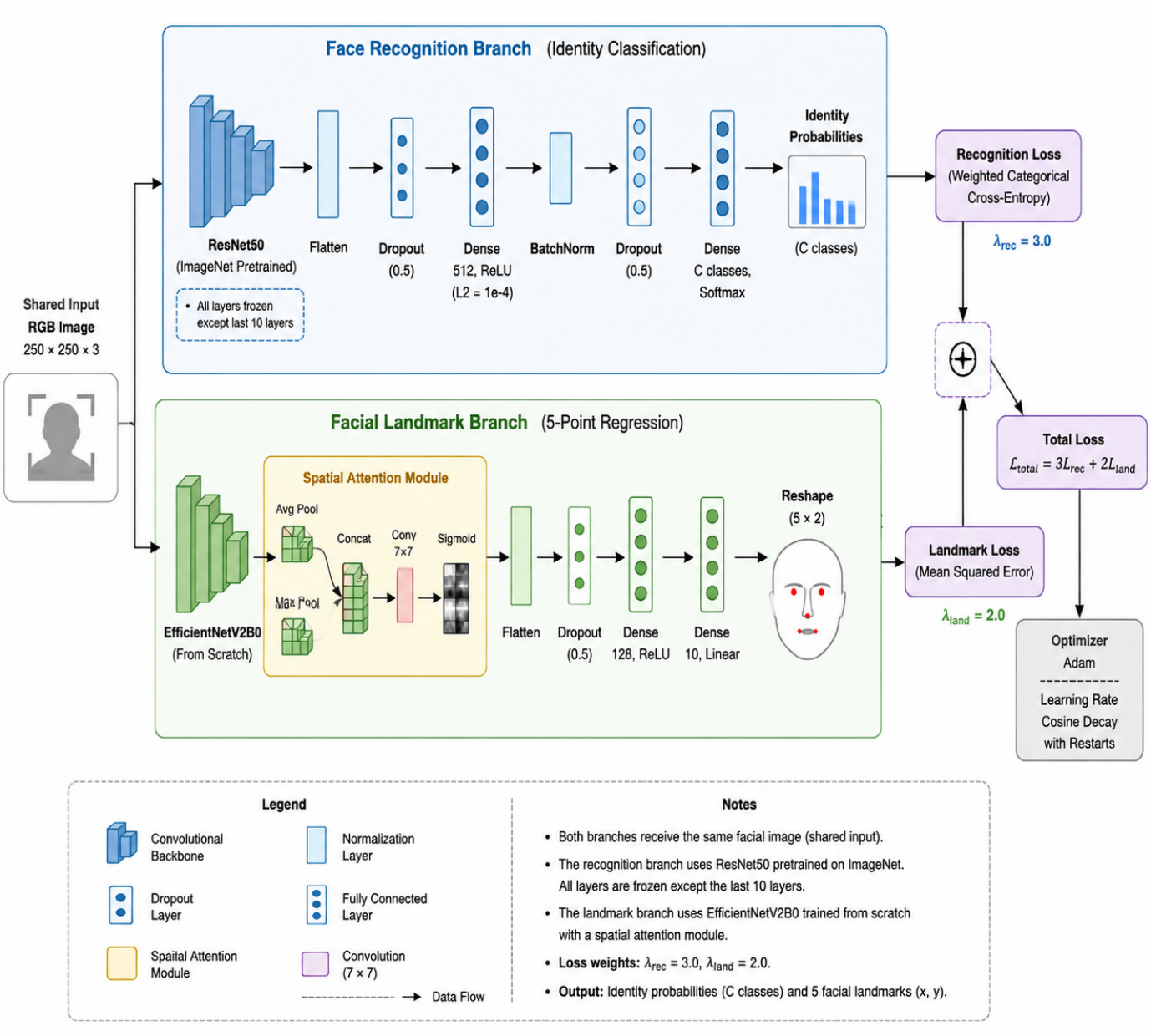}
  \caption{Architecture of the proposed dual-branch multi-task framework for
  pose-invariant face recognition and facial landmark localization. The ResNet50
  branch performs identity classification, while the EfficientNetV2B0 branch performs
  five-point landmark regression using spatial attention.}
  \label{fig:architecture}
  \end{figure*}

  Fig.~\ref{fig:architecture} illustrates the overall architecture.
  
The recognition branch is based on ResNet50. Residual networks enable effective training of deep architectures by introducing shortcut connections that reduce optimization difficulties associated with increasing network depth \cite{he2016resnet}.
  
The ResNet50 model is initialized with ImageNet weights. During training, all layers except the final ten layers are frozen, allowing higher-level features to adapt to the identity classification task.
  
The extracted representation is flattened and passed through dropout with a rate of 0.5, followed by a fully connected layer containing 512 rectified linear unit (ReLU)-activated units with $L_2$ regularization of $10^{-4}$. Batch normalization and an additional dropout layer are applied before the final softmax classifier.
  
  For $C$ identities, the recognition output is defined as:
  
  \begin{equation}
  \hat{Y} =
  [\hat{y}_{1},\hat{y}_{2},...,\hat{y}_{C}],
  \end{equation}
  
  where $\hat{y}_{c}$ represents the predicted probability of identity $c$.

The landmark localization branch employs EfficientNetV2B0, which provides an accuracy-efficiency trade-off through improved scaling and training strategies \cite{tan2021efficientnetv2}.
  
The extracted feature maps are refined using spatial attention and subsequently passed through dropout with a rate of 0.5 and a fully connected layer containing 128 ReLU units. The final linear layer predicts ten values corresponding to five two-dimensional facial landmarks:
  
  \begin{equation}
  (\hat{x}_{1},\hat{y}_{1}),...,
  (\hat{x}_{5},\hat{y}_{5}).
  \end{equation}

  Here, $(\hat{x}_{k},\hat{y}_{k})$ denotes the predicted pixel coordinates of
  landmark $k$, for $k=1,\ldots,5$.

  \subsection{Spatial Attention Mechanism}
  \label{subsec:attention}
  
  Sometimes pose variation and partial coverage can make some face regions less visible or give them less weight by lowering their importance. Spatial attention mechanism, in this proposed study, for landmark branch highlights key facial features independent from pose or occlusions.

  For this research work it was used spatial attention approach from CBAM \cite{woo2018cbam}. Across the feature map $F$ it was applied average pooling and max pooling, then two feature maps are concatenated and processed using a $7 \times 7$ convolution with sigmoid  activation:
  
  \begin{equation}
  M_s(F)=
  \sigma
  (
  f^{7\times7}
  [
  AvgPool(F);
  MaxPool(F)
  ]
  ),
  \end{equation}
  
  where $\sigma$ represents sigmoid activation.
  
  The refined feature representation is obtained as:
  
  \begin{equation}
  F'=F\odot M_s(F),
  \end{equation}
  
  where $\odot$ denotes element-wise multiplication.

  \subsection{Loss Functions and Multi-Task Optimization}
  \label{subsec:loss}

  The proposed framework combines two objectives: identity classification and facial
  landmark localization.
  
  For identity recognition, weighted categorical cross-entropy is used:
  
  \begin{equation}
  L_{rec}
  =
  -\frac{1}{N}
  \sum_{i=1}^{N}
  w_{c_i}
  \sum_{c=1}^{C}
  y_{ic}\log(\hat{y}_{ic}),
  \end{equation}

where $N$ is the number of samples in a batch, $C$ is the number of identities, $c_i$ is the ground-truth class index for sample $i$, $w_{c_i}$ is the weight assigned to class $c_i$, $y_{ic}$ is the one-hot encoded ground-truth indicator, and $\hat{y}_{ic}$ is the predicted probability that sample $i$ belongs to class $c$.

  Facial landmark localization is optimized using mean squared error:
  
  \begin{equation}
  L_{land}
  =
  \frac{1}{2KN}
  \sum_{i=1}^{N}
  \sum_{k=1}^{K}
  [
  (x_{ik}-\hat{x}_{ik})^2+
  (y_{ik}-\hat{y}_{ik})^2
  ],
  \end{equation}
  
where $K=5$ represents the number of facial landmarks, $(x_{ik},y_{ik})$ denotes the ground-truth pixel coordinates of landmark $k$ for sample $i$, and $(\hat{x}_{ik},\hat{y}_{ik})$ denotes the corresponding predicted coordinates. Because the coordinates are expressed in pixels, the landmark MSE is reported in squared pixels ($\mathrm{pixel}^2$).

  The final optimization objective is:
  
  \begin{equation}
  L=
  \lambda_{rec}L_{rec}
  +
  \lambda_{land}L_{land}.
  \end{equation}

  In the experiments:
  
  \begin{equation}
  \lambda_{rec}=3.0,\qquad
  \lambda_{land}=2.0.
  \end{equation}

  These weights were selected empirically during preliminary experiments to balance the different numerical scales of classification and regression losses.

  \subsection{Dataset and Data Preparation}
  \label{subsec:dataset}
  
  This study uses an adapted version of the Cross-Pose Labeled Faces in the Wild (CPLFW) dataset, which is designed to evaluate face recognition systems under large pose variations\cite{zheng2018cplfw}.

  Initially, this dataset was designed for face recognition and does not contain the necessary data structure for combined landmark localization. To address this issue dataset was restructured and for each image it was added identity label and five landmark points. The resulting dataset was more suitable for both recognition and landmark prediction during training. One of the important parts of face recognition systems is dataset preparation, so this aspect was studied separately. Research related to the pipeline to improve face recognition datasets used  a combination of face alignment, learned embeddings, and clustering to create or clean face datasets \cite{gallo2018pipeline}. This work helps with dataset quality, while this study focuses on preparing identity and landmark targets for joint training.
 
  Each image was resized to $250\times250$ pixels and converted into RGB format. The
  prepared dataset consists of:
  
  \begin{itemize}
  \item facial images with shape $(N,250,250,3)$;
  \item landmark coordinates with shape $(N,5,2)$;
  \item bounding-box coordinates with shape $(N,4)$;
  \item one-hot encoded identity labels with shape $(N,C)$.
  \end{itemize}

  \begin{figure*}[t]
  \centering
  \includegraphics[width=0.68\textwidth]{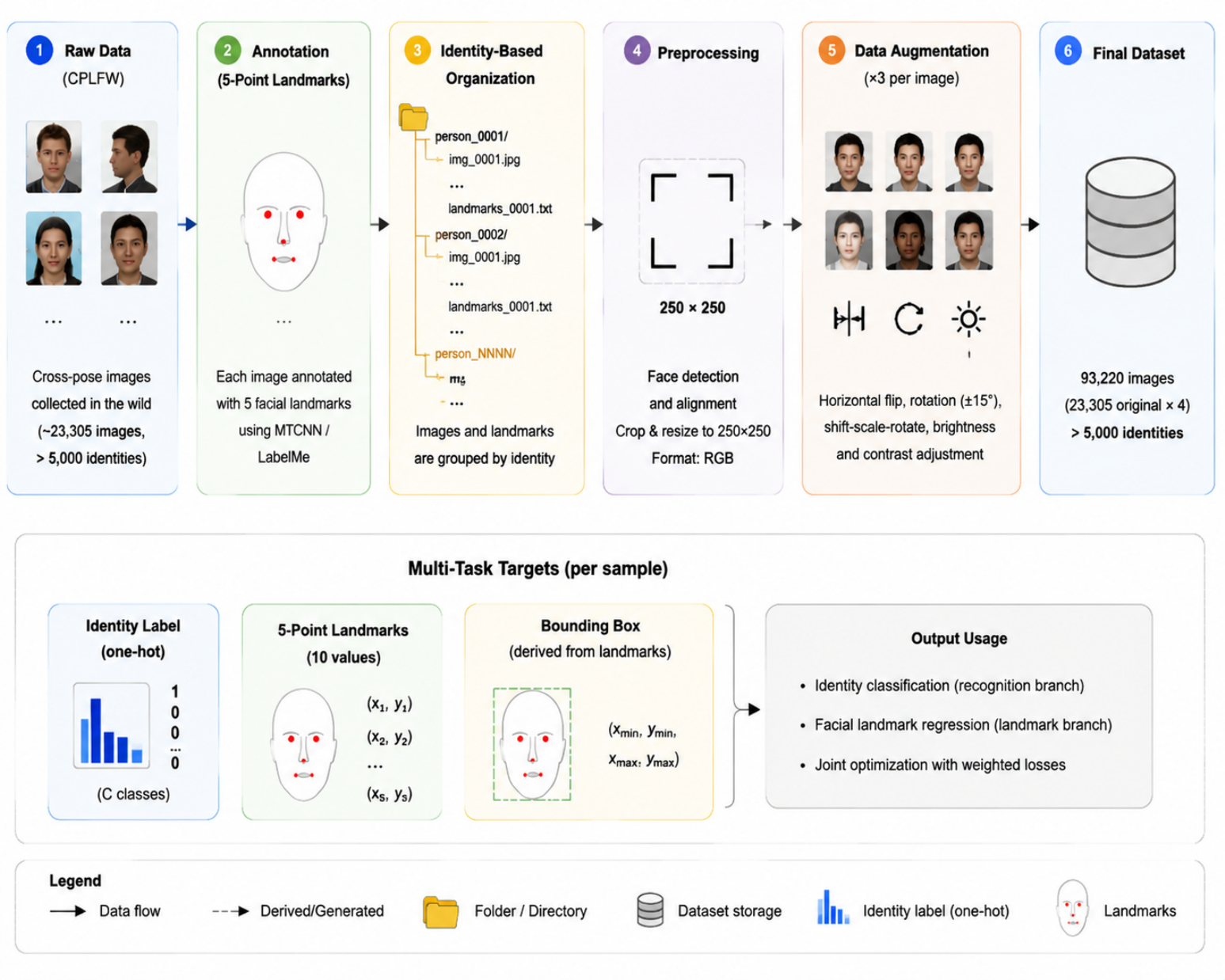}
  \caption{Preparation pipeline of the adapted CPLFW dataset for multi-task learning.}
  \label{fig:dataset}
  \end{figure*}

  \subsection{Data Augmentation}
  \label{subsec:augmentation}

Data augmentation was applied to improve model generalization. The augmentation strategy included horizontal flipping with probability 0.5, random rotations within approximately $\pm15^\circ$, shift-scale-rotate transformations, and brightness and contrast adjustments.
  
Three augmented samples were generated for each original image, increasing the dataset size from 23,305 original images to 93,220 samples.

  \subsection{Training and Implementation}
  \label{subsec:training}
  
  This research work uses Python 3.9 with TensorFlow 2.x and Keras. For training process it was applied a batch size of 16 for a maximum of 40 epochs. Preserving the class distribution the dataset was divided into train and validation sets, where 20\% of samples from each identity were for validation. 

  For optimization it was used Adam optimizer \cite{kingma2015adam}, while we set initial learning rate to $10^{-4}$ and regulated by cosine-decay-with-restarts scheduling. Moreover, to improve the generalization dropout, $L_2$ regularization, and batch normalization were applied. If during 10 epochs the validation performance did not improve it was stopped and the weights from the best epoch were retained. This process is early stopping.
  
  The final single deep learning model contains task-specific feature extraction, spatial attention,   identity classification, landmark regression, and weighted multi-task optimization and is presented as one framework.

\section{Results}
\label{sec:results}

\subsection{Experimental Setup}
\label{subsec:experimental_setup}
In this research work for experiments it was used Python 3.9 with TensorFlow 2.x and Keras frameworks, while CPLFW dataset with training strategy described in Section~\ref{subsec:training} was used to evaluate multi-task framework. Classification accuracy determined face recognition performance and mean squared error (MSE), measured in squared pixels ($\mathrm{pixel}^2$), between predicted and ground-truth landmark coordinates evaluated facial landmark localization.

Model was trained by using a batch size of 16 for a maximum of 40 epochs and Adam optimizer  with an initial learning rate of $10^{-4}$. Early stopping with a patience of 10 epochs was required to avoid overfitting and restore the best-performing model parameters. Also, a cosine decay with restarts learning-rate scheduler helped the training process converge smoothly.

\subsection{Training Performance Analysis}
\label{subsec:training_results}

\begin{figure}[t]
\centering
\includegraphics[width=\columnwidth]{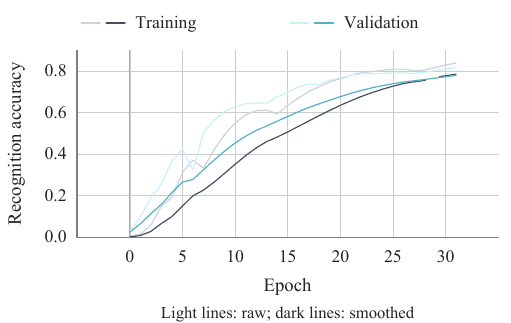}
\caption{Training and validation accuracy of the face recognition branch during
optimization.}
\label{fig:accuracy_curve}
\end{figure}

\begin{figure}[t]
\centering
\includegraphics[width=\columnwidth]{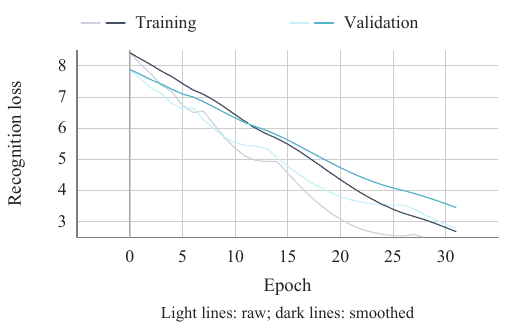}
\caption{Training and validation loss of the face recognition branch.}
\label{fig:recognition_loss}
\end{figure}

\begin{figure}[t]
\centering
\includegraphics[width=\columnwidth]{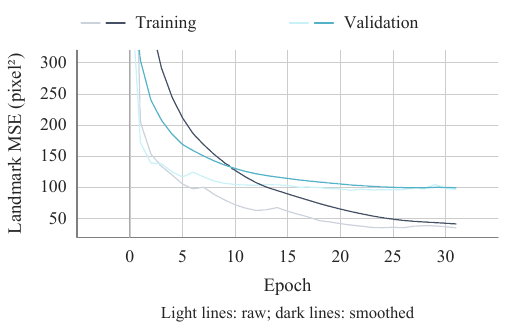}
\caption{Training and validation mean squared error for facial landmark
localization.}
\label{fig:landmark_mse}
\end{figure}

\begin{figure}[t]
\centering
\includegraphics[width=\columnwidth]{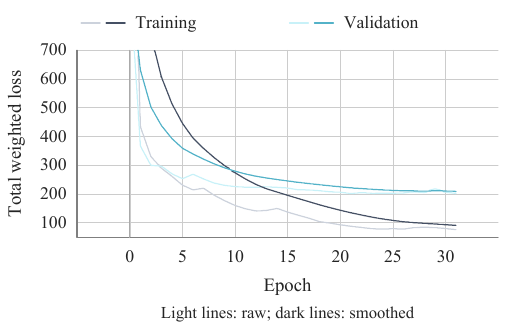}
\caption{Overall training and validation loss of the proposed multi-task framework.}
\label{fig:total_loss}
\end{figure}

The training dynamics of the proposed multi-task framework are presented in Figs.~\ref{fig:accuracy_curve}--\ref{fig:total_loss}. The recognition branch demonstrated stable improvement throughout the optimization process. As shown in Fig.~\ref{fig:accuracy_curve}, the training accuracy increased to 83.87\%, while the validation accuracy reached 81.58\% at the final training stage.

The small difference between training and validation accuracy indicates that the model maintained reasonable generalization capability despite the complexity of the cross-pose recognition task. The applied regularization techniques, including dropout, weight decay, and early stopping, contributed to reducing excessive overfitting.

Fig.~\ref{fig:recognition_loss} illustrates the recognition loss evolution. The classification loss continuously decreased during training, demonstrating effective adaptation of the ResNet50-based recognition branch to identity-specific features. The final validation recognition loss was 2.77.

The landmark localization performance is shown in Fig.~\ref{fig:landmark_mse}. The localization error decreased substantially during the first epochs and gradually converged as training progressed. The final training MSE was 34.73~$\mathrm{pixel}^2$, while the validation MSE reached 96.38~$\mathrm{pixel}^2$.

The larger difference between training and validation MSE compared with the classification task is expected because landmark regression is more sensitive to annotation noise, pose variations, and small spatial errors in facial key-point coordinates.

Fig.~\ref{fig:total_loss} presents the overall optimization objective combining identity recognition and landmark localization losses. The total loss decreased from the initial high values and gradually converged during training. The final training loss was 75.46, while the validation loss reached 201.36.

The convergence of the joint objective demonstrates that simultaneous optimization of classification and regression objectives did not negatively affect either task. Instead, the multi-task strategy enabled the model to learn complementary semantic and spatial facial representations.

\subsection{Quantitative Results}
\label{subsec:quantitative_results}

Summary of this research study is illustrated in Table~\ref{tab:quantitative_results} and according to these data, the model recorded a validation accuracy of 81.58\% for identity recognition and a validation MSE of 96.38~$\mathrm{pixel}^2$ for facial landmark localization.

These results show that proposed multi-task architecture effectively covers both identity classification and facial structure estimation  across pose variations. Effective balance between recognizing faces and understanding their geometric features was achieved by combination of task-specific backbones with weighted multi-task optimization.

\begin{table}[t]
\centering
\caption{Quantitative evaluation of the proposed multi-task framework.}
\label{tab:quantitative_results}
\begin{tabular}{lc}
\hline
Metric & Value \\
\hline
Training Recognition Accuracy & 83.87\% \\
Validation Recognition Accuracy & 81.58\% \\
Training Landmark MSE ($\mathrm{pixel}^2$) & 34.73 \\
Validation Landmark MSE ($\mathrm{pixel}^2$) & 96.38 \\
\hline
\end{tabular}
\end{table}

\subsection{Qualitative Analysis}
\label{subsec:qualitative_results}

\begin{figure}[t]
\centering
\includegraphics[width=\columnwidth]{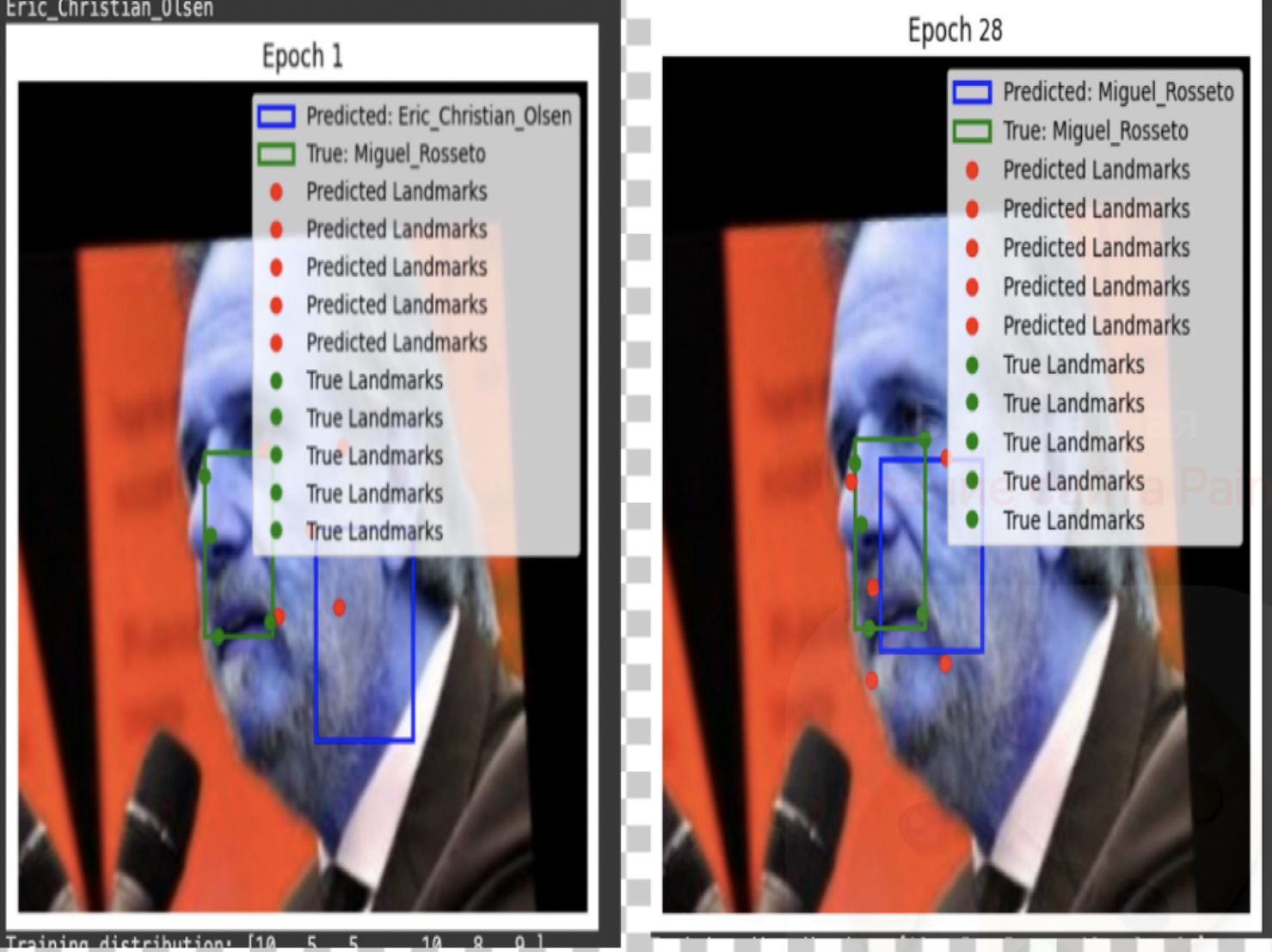}
\caption{Qualitative visualization of identity prediction and facial landmark
localization during training.}
\label{fig:qualitative}
\end{figure}

Fig.~\ref{fig:qualitative} provides a qualitative visualization of the model behavior during training. The visualization demonstrates the simultaneous learning of identity recognition and facial landmark localization.

Despite the big difference of predicted identity and landmark position from the ground-truth values at early training steps, after applying optimizations, the model improved face recognition and produced more accurate facial landmark localization. This shows how our MTL framework can effectively learn high-level identity features and spatial facial structures.

\section{Discussion}
\label{sec:discussion}

\subsection{Effectiveness of the Multi-Task Framework}
\label{subsec:mtl_discussion}

Traditional face recognition methods usually focus on identifying people, while in this approach adding a landmark localization allows to consider facial structure. The results of this study illustrate how well the proposed framework works in combining identity recognition and facial landmark localization into single model. Training of two branches in one framework develops facial representations that have both identity details and structural information, where recognition branch responsible for capturing the high-level features needed to tell faces apart and landmark branch understanding the spatial relationships among key facial features. The success of this method can be proved by validation accuracy of 81.58\% and landmark localization MSE of 96.38~$\mathrm{pixel}^2$. These metrics mean that model can effectively handle both tasks even when faces are viewed from different angles. The findings of this study show that landmark localization and identity recognition can be trained together within the proposed dual-branch framework.

\subsection{Role of Spatial Attention}
\label{subsec:attention_discussion}

Pose variation introduces significant changes in facial appearance, causing some facial regions to become less informative depending on head orientation. To address this challenge, the proposed landmark localization branch incorporates a spatial attention mechanism to emphasize important facial regions.

This approach supports the idea that key facial features like the eyes, nose, and mouth are important for accurately locating landmarks, therefore attention module helps the model improve its understanding of the image by giving different levels of importance to various spatial locations.

\subsection{Analysis of Training Behaviour}
\label{subsec:training_discussion}

The training progress shows that our multi-task framework is learning steadily and reliably. We see a steady increase in recognition accuracy as training goes on, while both recognition loss and landmark localization error gradually decrease. Notably, the difference between training and validation performance is more apparent for the landmark localization task than for identity recognition. This is because landmark regression is more sensitive to factors like annotation quality, pose variation, and small coordinate shifts. On the other hand, identity classification depends on higher-level facial features that tend to be more resilient to minor spatial differences. Incorporating techniques like dropout, $L_2$ regularization, early stopping, and learning-rate scheduling has helped us prevent overfitting and ensure a more stable training process.

\subsection{Comparison with Existing Approaches}
\label{subsec:comparison}

Previous facial analysis studies have commonly focused on individual tasks, such as face recognition, landmark localization, or attribute prediction. DeepFace, FaceNet, and ArcFace demonstrated the effectiveness of deep representation\cite{taigman2014deepface,schroff2015facenet,deng2019arcface} learning for face recognition, while HyperFace showed the potential of multi-task learning for combining several facial analysis objectives\cite{ranjan2017hyperface}.

The difference of proposed framework is it combines face recognition and landmark localization into one model, instead of preprocessing landmark localization as independent step. The results of this study are based on the evaluation of the model on the CPLFW dataset.

\subsection{Limitations and Future Work}
\label{subsec:limitations}

While the initial results are promising, there is still room for growth. Right now, the study mainly uses the CPLFW dataset, which centers on cross-pose face recognition. Exploring larger datasets with a variety of lighting conditions, occlusions, and diverse demographics would really help solidify the findings. Additionally, future studies could include ablation tests to understand how each part, like the spatial attention module or multi-task goals, contributes to the overall performance. Looking ahead, it would be exciting to see research on adaptive loss-weights, bigger datasets, and cutting-edge architectures to make facial analysis more reliable in real-world situations. Validating the model on broader benchmarks is a good direction for the future development of the provided method. One of the benchmarks is the multi-face dataset prepared by Dubey and Mukherjee \cite{dubey2018multiface}. The cross-spectral FaceSurv dataset gives an opportunity for testing face detection under varying resolutions and modalities\cite{gupta2019facesurv}. It is also interesting to explore synthetic training data tools like SegTex \cite{ambardi2023segtex}. This tool generates textured face images from segmentation maps, which can improve data diversity. These works require the development of evaluation protocols and annotations, but exploring these approaches can help improve the performance of the provided model.

\section{Conclusion}
\label{sec:conclusion}

In this study, a multi-task deep learning framework was introduced for two primary purposes: pose-invariant face recognition and facial landmark localization, where ResNet50 was used for identity recognition branch and EﬀicientNetV2B0 with spatial attention for landmark localization branch.

The findings illustrate how combination of identity classification and facial landmark regression can be delivered in single model, where two tasks can help each other to learn better. Moreover, we can see how attention-based multi-task architectures handle real world  and unconstrained environment challenges during face recognition.

Provided framework have strong ability to handle both recognition and localization tasks during pose invariance, it can be proven by achieved a validation recognition accuracy of 81.58\% and a landmark localization MSE of 96.38~$\mathrm{pixel}^2$ on CPLFW dataset.

In future work we are aiming to train our model on larger and more diverse dataset, provide more detailed ablation to show how different components contribute to results. One of the possible explorations can be connected to optimization of multi-task learning by using adaptive loss-weighting approaches.

\IEEEtriggeratref{36}
\bibliographystyle{IEEEtran}
\bibliography{export}
\end{document}